\documentclass[twocolumn,twoside,journal]{IEEEtran}

\usepackage{amsmath}
\usepackage{color}
\usepackage{amssymb}
\usepackage{amsthm}
\usepackage{graphicx}
\usepackage{multirow}
\usepackage{balance}
\usepackage{caption}
\usepackage{algorithm,algorithmic}

\begin{document}
\bibliographystyle{IEEEtran}
\title{Laplacian Prior Variational Automatic Relevance Determination for Transmission Tomography}

\author{\IEEEauthorblockN{Jingwei Lu, David G. Politte,~\IEEEmembership{Member,~IEEE}, Joseph A. O'Sullivan,~\IEEEmembership{Fellow,~IEEE}}

\thanks{This work was supported in part by the National Institutes of Health under research grant R01CA212638.\

 Jingwei Lu and Joseph A. O'Sullivan are with the Department of Electrical and System Engineering, Washington University in St. Louis, St. Louis,
MO, 63130 (email: jlu23@wustl.edu,  jao@wustl.edu)

David G. Politte is with Mallinckrodt Institute of Radiology, Washington University in St. Louis School of Medicine, St. Louis, MO, 63130 (email: politted@wustl.edu) }
\thanks{}}

\markboth{}%
{Lu \MakeLowercase{\textit{et al.}}: Laplacian Prior Variational Automatic Relevance Determination for Transmission Tomography }

\maketitle

\begin{abstract}
In the classic sparsity-driven problems, the fundamental L-1 penalty method has been shown to have good performance in reconstructing signals for a wide range of problems. However this performance relies on a good choice of penalty weight which is often found from empirical experiments. We propose an algorithm called the Laplacian variational automatic relevance determination (Lap-VARD) that takes this penalty weight as a parameter of a prior Laplace distribution. Optimization of this parameter using an automatic relevance determination framework results in a balance between the sparsity and accuracy of signal reconstruction. Our algorithm is implemented in a transmission tomography model with sparsity constraint in wavelet domain.
\end{abstract}


%
\IEEEpeerreviewmaketitle

\section{Introduction}
\IEEEPARstart{T}{ransmission} tomography is a well-established and efficient method to represent the linear attenuation coefficient inside the image of interest. Among all the modalities, X-ray computed tomography (CT) \cite{Kalender2006} introduced into clinical use in 1972 is the earliest and also the most widely used one. The pursuit to get high resolution requires a large amount of measurement data; however, limitations of data storage and computation resources force a threshold on the resolution. To overcome these limitations, extra information of the image itself should be taken into consideration. The sparsity of image can be represented not only in the image domain, but also in a transformed domain like the directional gradient domain or wavelet transform domain  \cite{Simoncelli1999}\cite{Xu2014}.

In sparsity-driven problems, the basic idea is that when we know in advance that the image of interest is sparse or sparse in some basis, then we need fewer observations than the traditional methods to  reconstruct the image using the significant components in the image \cite{Baraniuk2007}. For these problems, a universal method to seek a best sparse approximation is an L-1 penalty 
$$L(x)=f(x)+\lambda|x|\eqno{(1.1)}$$ 
where $L(x)$ is the cost function, and $f(x)$ is the data fitting term. This is equivalent to the L-1 regularization problem \cite{Candes2006}
$$\min |x| \ s.t.\   f(x)\leq \epsilon\eqno{(1.2)}$$
The problem to minimize (1.1) is called LASSO \cite{Tibshirani1996}. The optimal solution shrinks to 0 as the parameter $\lambda$ goes to infinity, which means that the sparsity in the reconstruction is determined by the choice of $\lambda$. A very small $\lambda$ yields low sparsity. However, if $\lambda$ is too large, it over-penalizes and we have an all-zero image in the limit \cite{Zou2006}. To get a balance between sparsity and data fitting, we propose the Lap-VARD algorithm to automatically update $\lambda$ iteratively. Because in this algorithm $\lambda$ depends on the measurements, it is an adaptive method. 

This Lap-VARD is inspired by the fact that if we take $f(x)=-\log{p(y|x)}$ in (1.1), and let the prior of $x$ be a zero-mean Laplace distribution: $p(x;\lambda)=\frac{1}{2\lambda}\exp(-\frac{|x|}{\lambda})$,with $\text{E}[x]=0$ and $\text{Var}[x]=2\lambda^2$, then $$L(x)=-\log{p(y|x)-\log{p(x;\lambda)}}.\eqno{(1.3)}$$
Now, this problem becomes a MAP problem \cite{Hebert1992}. Instead of maximizing over $x$, we take $\lambda$ as hyper-parameter and try to maximize the marginal likelihood $p(y;\lambda)$
$$\max\limits_\lambda L(\lambda)=\max\limits_\lambda\log p(y;\lambda)=\max\limits_\lambda\log\int p(y|x)p(x;\lambda)dx.\eqno{(1.4)}$$
This method is called automatic relevance determination (ARD) \cite{Wipf2008}. In  \cite{Kaganovsky2014}, the AM-ARD method is taken into the Transmission Tomography model with a Gaussian Prior. In the Lap-VARD algorithm, we use the Laplace Prior to promote the sparsity and automatically get the best-fitting hyper-parameter $\lambda$ in (1.1) to have a balance between sparsity and data-fitting.

\section{Proposed Method}
In X-ray CT image reconstruction, the data fitting term is the negative log-likelihood of a Poisson distribution \cite{Fessler2000}
$$f(u)=-\log p(y|u)=\sum_i[-y_i\log q_i(u)+q_i(u)],\eqno{(2.1)}$$
where $u$ is the image we try to reconstruct, $y_i$ is the X-ray photon measurement for source-detector pair $i$, and $q_i(u)=I_i\exp(-\sum_j h_{ij}u_j)$ is the mean 
of a Poisson random variable $y_i$, $I_i$ is the air scan photon counts for source-detector pair $i$, and $h_{ij}$ is an element in the system matrix that represents the contribution of pixel $j$ to source-detector pair $i$ \cite{DeMan2001a}.

In \cite{Jingwei}, a wavelet sparsity penalty is introduced and the advantage of a wavelet sparsity penalty is that it does not generate biased reconstructions. The wavelet sparsity penalty is in the following form
$$\sum_k g(\beta_k;\gamma_k)=\sum_k\gamma_k |\beta_k|, \eqno{(2.2)}$$
where $g(\beta_k;\gamma_k)$ is the negative log-likelihood of a Laplace distribution with $0$ mean and variance $\frac{2}{\gamma_k^2}$. $\beta$ is the set of wavelet coefficients of image $u$ with the transform pair
$$u=\Omega \times \beta \eqno{(2.3)}$$
and $$\beta=\hat{\Omega}\times u, \eqno{(2.4)}$$
where $\Omega\in R^{K\times J}$ and $\hat{\Omega}\in R^{J\times K}$. From (1.1), the overall cost function is 
$$L(u)=-\log p(y|u)+\sum_k \gamma_k  |\beta_k|, \eqno{(2.5)}$$
where
$$\beta=\hat{\Omega}\times u.$$
Reformulating the cost function in the wavelet coefficient domain only, the reformulated cost function is 
\begin{align*}
L(\beta)=&-\log p(y|{\Omega}\beta)+\sum_k \gamma_k  |\beta_k|\\
        =&\sum_i[-y_i\log q_i(\beta)+q_i(\beta)]+\sum_k \gamma_k  |\beta_k|, \tag{2.6}  
\end{align*}
where 
$$q_i(\beta)=I_i \exp(-\sum_j h_{ij}\sum_k \omega_{kj}\beta_k)=I_i \exp(-\sum_k \phi_{ik}\beta_k).\eqno{(2.7)}$$
Now, we have a new system matrix $\Phi = H\times \Omega.$ The performance of the reconstruction algorithm is controlled by the hyper-parameters $\gamma$. We use the ARD framework to find the optimal $\gamma$.

The marginal log-likelihood $\log p(y;\gamma)$ is 
$$\log p(y;\gamma)=\log \int{p(y|\Omega\beta)p(\beta;\gamma)d\beta}.\eqno{(2.8)}$$ 
Then $\gamma^*=\arg\max\limits_\gamma \log p(y;\gamma)$. Here, we rewrite the marginal log-likelihood $\log p(y|\gamma)$ as
\begin{align*}
\log p(y|\gamma)=&-E_{q(\beta)}\log[q(\beta)/p(y,\beta;\gamma)]\\
                 &+D_{KL}[q(\beta)||p(\beta|y;\gamma)] \tag{2.9}
\end{align*}
where $E_{q(\beta)}$ stands for the expected value with respect to $q(\beta)$, and $D_{KL}$ is the Kullback-Leibler (KL) divergence.
 
From \cite{Neal1998}, we take a variational method to solve $(2.9)$ iteratively. Since the change of $q(\beta)$ does not change the value of $\log p(y|\gamma)$, then at iteration $t$ we set $q(\beta)^{(t+1)}=p(\beta|y,\gamma^{(t)})$ such that $D_{KL}[q(\beta)||p(\beta|y;\gamma)]=0$. we just need to maximize the first term in $(2.9)$ with respect to $\gamma$, which is called \emph{free variational energy} (FVE). The EM algorithm \cite{Neal1998} can be viewed as minimizing the FVE function by alternating between updating $q(\beta)$ and $\gamma$. However, the expression for $p(\beta|y,\gamma^{(t)})$ is complicated. As in VARD, we restrict the form of the posterior distribution. Here, we still use Laplace distributions $q(\beta)\sim Laplace(\mu, b)$ with $\text{E}[\beta]=\mu$ and $\text{Var}[\beta]=2b^2$.

Then, the object function can be written as
\begin{align*}
F(\gamma,\mu,b)=&E_{q(\beta)}[-\log p(y|\Omega,\beta)]+D_{KL}[q(\beta)||p(\beta;\gamma)]\\ 
=&-\int q(\beta)\log p(y|\Omega,\beta)d\beta+\int q(\beta)\log q(\beta)d\beta\\
&-\int q(\beta)\log p(\beta;\gamma)d\beta\\
=&\sum\limits_i(\prod\limits_k\frac{1}{1-(b_k\Phi_{ik})^2})I_{i}\exp(-\sum\limits_k  \Phi_{ik}\mu_k)\\
&+\sum\limits_iy_i(\sum\limits_k\Phi_{ik}\beta_k)+\sum\limits_j\frac{1}{\gamma_j}(b_j\exp(-\frac{|\mu_j|}{b_j})+|\mu_j|)\\
&+\sum\limits_k\log (2\gamma_k)-\sum\limits_k\log(2b_k).\tag{2.10}
\end{align*}
Function $F$ is convex with respect to $\mu$ for fixed $b$ and convex with respect to $b$ for fixed $\mu$. Though $F$ is not convex with respect to $\gamma$, there only exists one stationary point for $\gamma$, and it is easy to show that $F(\gamma)$ has a global minimum. We can iteratively update $\mu$, $b$ and $\gamma$.
\paragraph{Optimization of $\gamma$}
$\gamma^*$ has an explicit expression $$\gamma^*_k=b_k\exp(-\frac{|\mu_k|}{b_k})+|\mu_k|.\eqno{(2.11)}$$ 

\paragraph{Optimization of $\mu$}
The terms in $F$ containing $\mu$ are\\
\begin{align*}
F(\mu)=&+\sum\limits_i(\prod\limits_k\frac{1}{1-(b_k\Phi_{ik})^2})I_{i}\exp(-\sum\limits_k\Phi_{ik}\mu_k)\\
&+\sum\limits_iy_i(\sum\limits_k\Phi_{ik}\mu_k)+\sum\limits_k\frac{1}{\gamma_k}(b_k\exp(-\frac{|\mu_k|}{b_k})+|\mu_k|).\tag{2.12}
\end{align*}
We decouple every $\mu_k$ using surrogate function
\begin{align*}
G(\mu)=&\sum\limits_kg(\mu_k)=\sum\limits_kb_k^y\mu_k+\theta^+_k\exp(-Z_1(\mu_k-\mu_k^{(t)}))\\
&-\theta^-_k\exp(Z_1(\mu_k-\mu_k^{(t)}))+\frac{b_k}{\gamma_k}e^{-\frac{|\mu_k|}{b_k}}+\frac{1}{\gamma_k}|\mu_k|, \tag{2.13}
\end{align*}
where \begin{align*} 
b^y=&\Phi'y \\
\theta^+_k=&\sum\limits_{i\in C}(\prod\limits_{k\in C}\frac{1}{1-(b_k\Phi_{ik})^2})\alpha_{ik}I_{i}\exp({-\sum\limits_k\Phi_{ik}\mu_k^{(t)}})\\
\theta^-_k=&\sum\limits_{i\in \hat{C}}(\prod\limits_{k\in \hat{C}}\frac{1}{1-(b_k\Phi_{ik})^2})\alpha_{ik}I_{i}\exp({-\sum\limits_k\Phi_{ik}\mu_k^{(t)}})\\
Z_1=&\frac{|\Phi_{ik}|}{\alpha_{ik}}=\max\limits_i\sum\limits_j|\Phi_{ik}|\\
C=&\{(i,k)|\phi_{ik}\geq 0\},\hat{C}=\{(i,k)|\phi_{ik}<0\}.
\end{align*}
Given the convex surrogate function with decoupled parameter $\mu$,  several methods are available. To simply account for the non-continuous term $|\mu_k|$, the sub-gradient method is chosen \cite{Boyd2006}.

\paragraph{Optimization of $b$} 
The terms in $F$ containing $b$ are 
\begin{align*}
F(b)=&\sum\limits_i(\prod\limits_k\frac{1}{1-(b_k\Phi_{ik})^2})I_{i}\exp({-\sum\limits_k\Phi_{ik}\mu_k})\\
&+\sum\limits_k\frac{1}{\gamma_k}(b_k\exp({-\frac{|\mu_k|}{b_k}}))-\sum\limits_k\log(2b_k).\tag{2.14}
\end{align*}
Just like the parameter $\mu$, the optimal $b$ does not have a closed-form solution. We still use Newton's method to optimize $b$. Since the parameters $b_k$ are coupled, we decouple every $b_k$ from the convex decomposition lemma \cite{OSullivan2007} 
$$f(x)\leq\sum\limits_j {r_j}f(\hat{x}+\frac{x_j-\hat{x_j}}{r_j}e_j).\eqno{(2.15)}$$
The final surrogate function for $F(b)$ is
\begin{align*}
G(b)=&\sum\limits_k g(b_k)=\sum\limits_k\frac{1}{\gamma_k}(b_k\exp({-\frac{|\mu_k|}{b_k}}))-\sum\limits_k\log(2b_k)\\
&+\sum\limits_i\sum\limits_kr_{ik} Q_{ik}(\hat b)\frac{1}{1-(\tilde{b_{k}}\Phi_{ik})^2}I_{i}\exp({-\sum\limits_k\Phi_{ik}\mu_k})\\
\tag{2.16}
\end{align*}
where
\begin{align*}
\sum\limits_k r_{ik}\leq& 1\\
Q_{ik}(\hat{b})=&\prod\limits_{j\neq k}\frac{1}{1-(\hat{b_j}\Phi_{ij})^2}\\
\tilde{b}_{ik}=&(\hat{b}_k+\frac{{b}_k-\hat{b}_k}{r_{ik}})
\end{align*}

\paragraph{Pseudo-code} The algorithm is summarized as:
\begin{algorithm}
\caption{Lap-VARD}
\label{alg1}
\begin{algorithmic}
\STATE Initialize $\mu$, $b$, $\gamma$
\FOR {$cont$=$1$ to $N$}
\STATE do backward projection, and compute $b^y$, $\theta^+_k$,$\theta^-_k$;
\STATE minimize (2.13) by Newton's method: $\hat{\mu}=\arg\min\limits_\mu G(\mu)$; 
\STATE do forward projection, and compute $Q_{ik}$, $\tilde{b_{ik}}$;
\STATE minimize (2.17) by Newton's method: $\hat{b}=\arg\min\limits_b G(b)$;
\STATE update $\gamma$ with $\gamma_k=b_k\exp({-\frac{|\mu_k|}{b_k}})+|\mu_k|$; 
\ENDFOR
\end{algorithmic}
\end{algorithm}

\begin{figure*}[!ht]
\begin{minipage}[!h]{0.25\textwidth}
\includegraphics[height=1.4in]{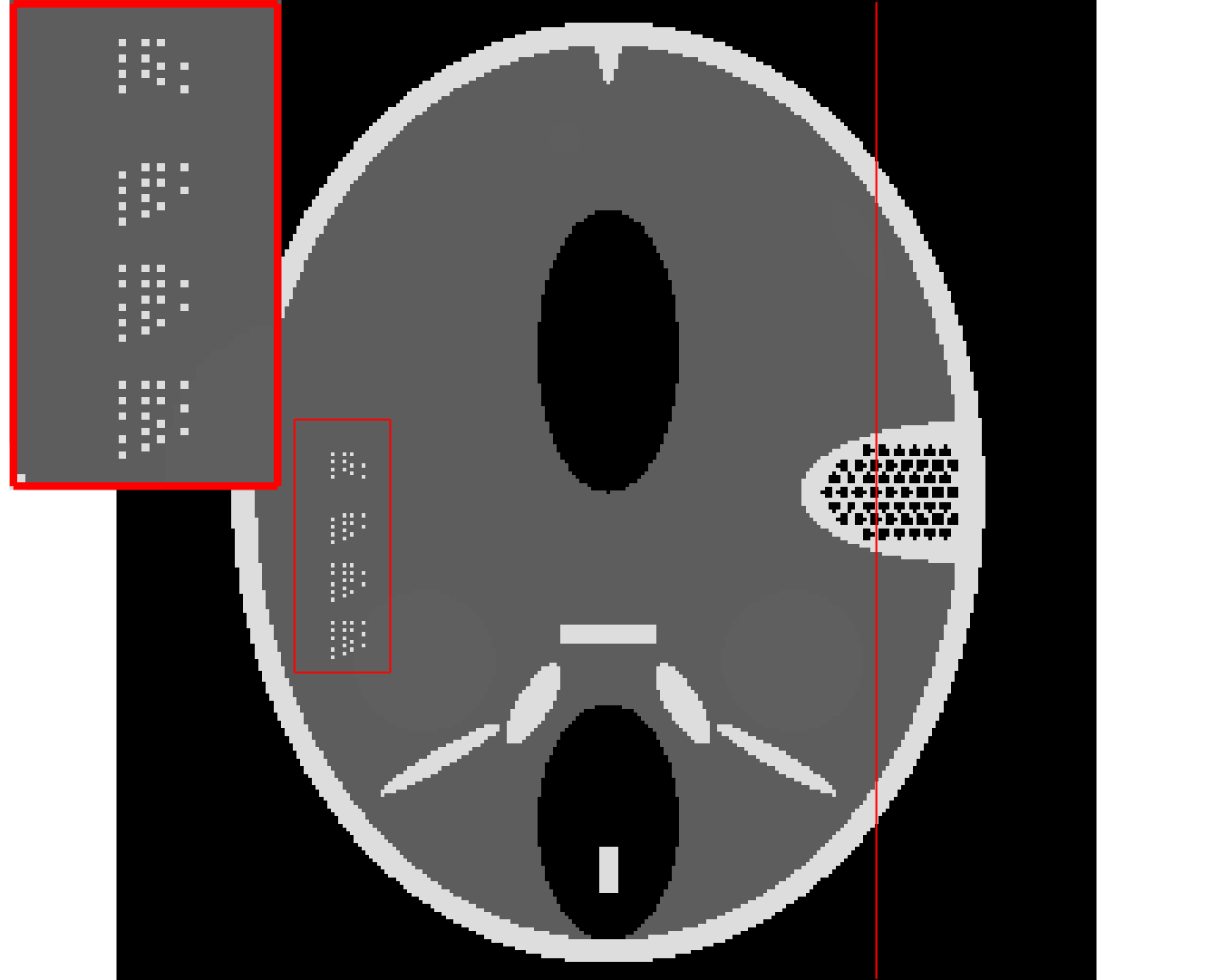}
\captionsetup{justification=centering}
\caption*{(a)}
\end{minipage}%
\begin{minipage}[!h]{0.25\textwidth}
\includegraphics[height=1.4in]{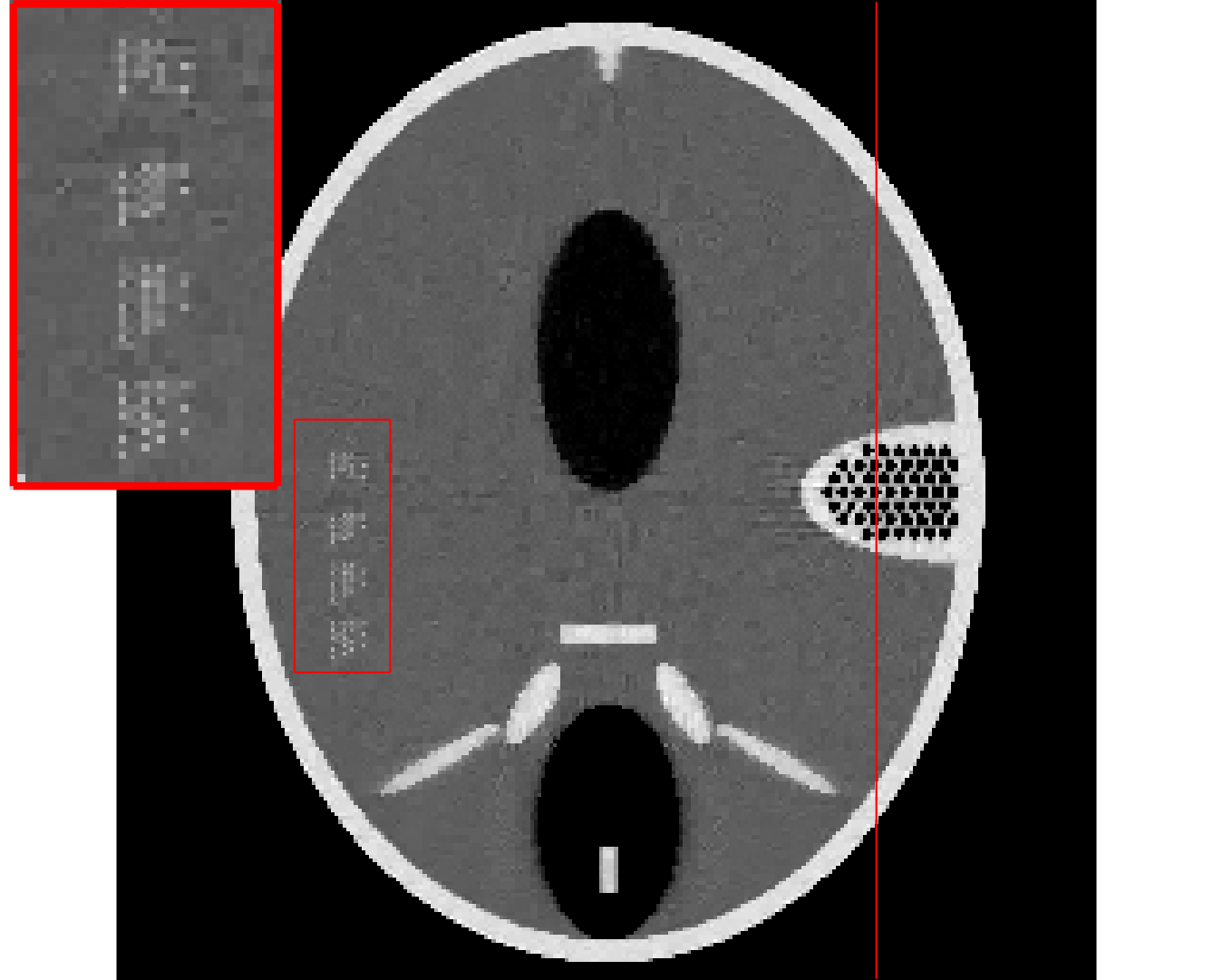}
\captionsetup{justification=centering}
\caption*{(b)}
\end{minipage}%
\begin{minipage}[!h]{0.25\textwidth}
\includegraphics[height=1.4in]{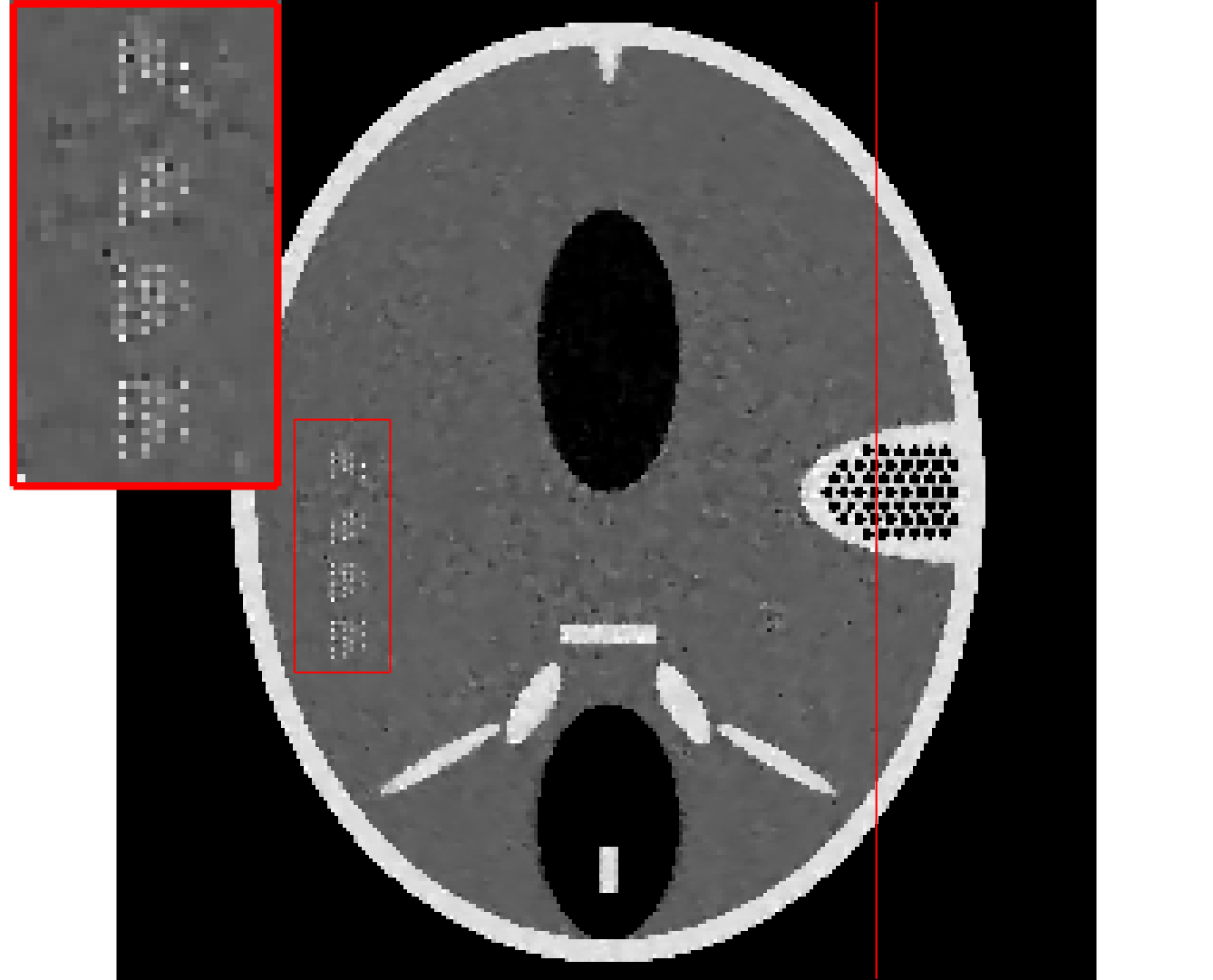}
\captionsetup{justification=centering}
\caption*{(c)}
\end{minipage}%
\begin{minipage}[!h]{0.25\textwidth}
\includegraphics[height=1.4in]{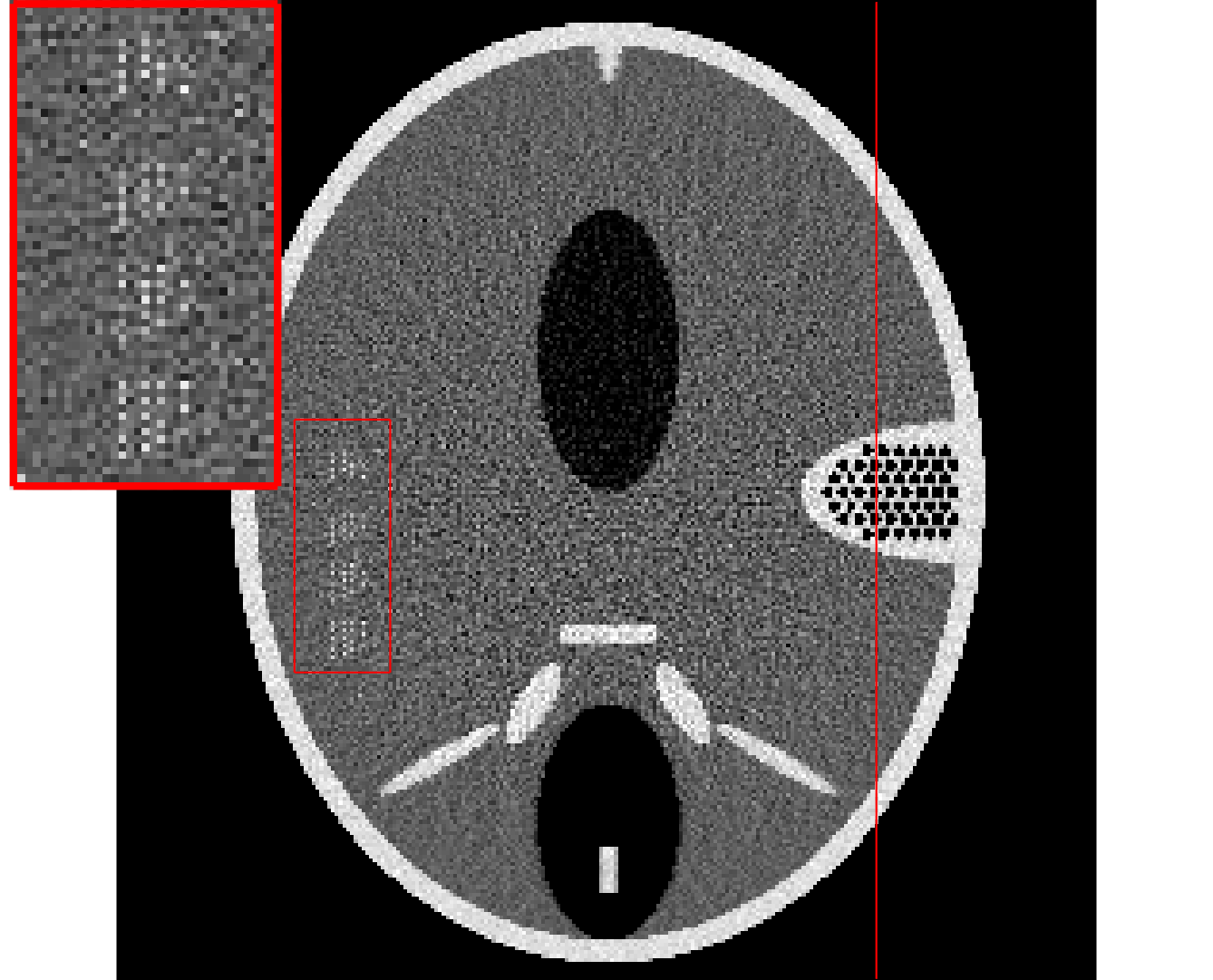}
\captionsetup{justification=centering}
\caption*{(d)}
\end{minipage}
\begin{minipage}[!h]{0.25\textwidth}
\includegraphics[height=1.4in]{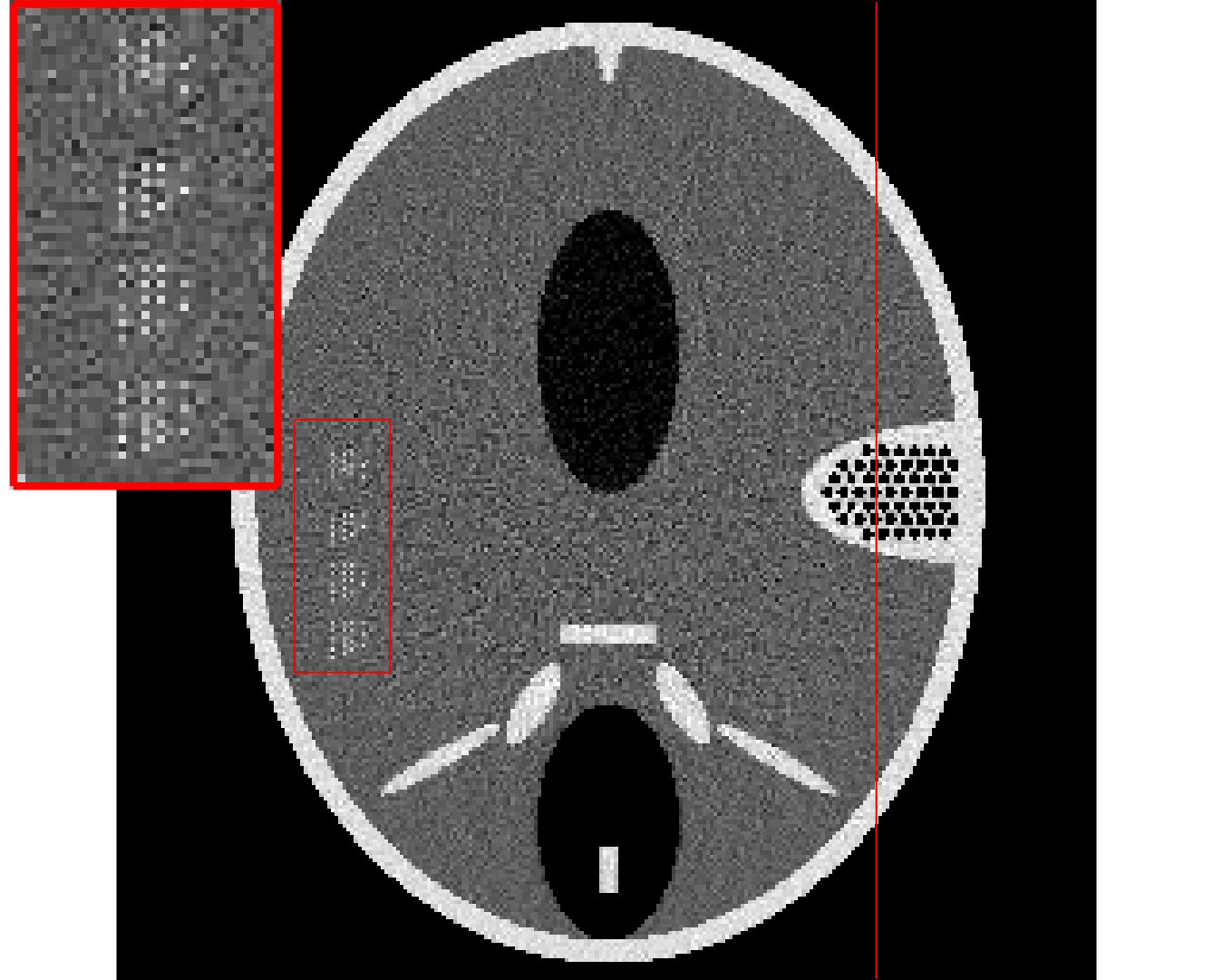}
\captionsetup{justification=centering}
\caption*{(e)}
\end{minipage}%
\begin{minipage}[!h]{0.25\textwidth}
\includegraphics[height=1.4in]{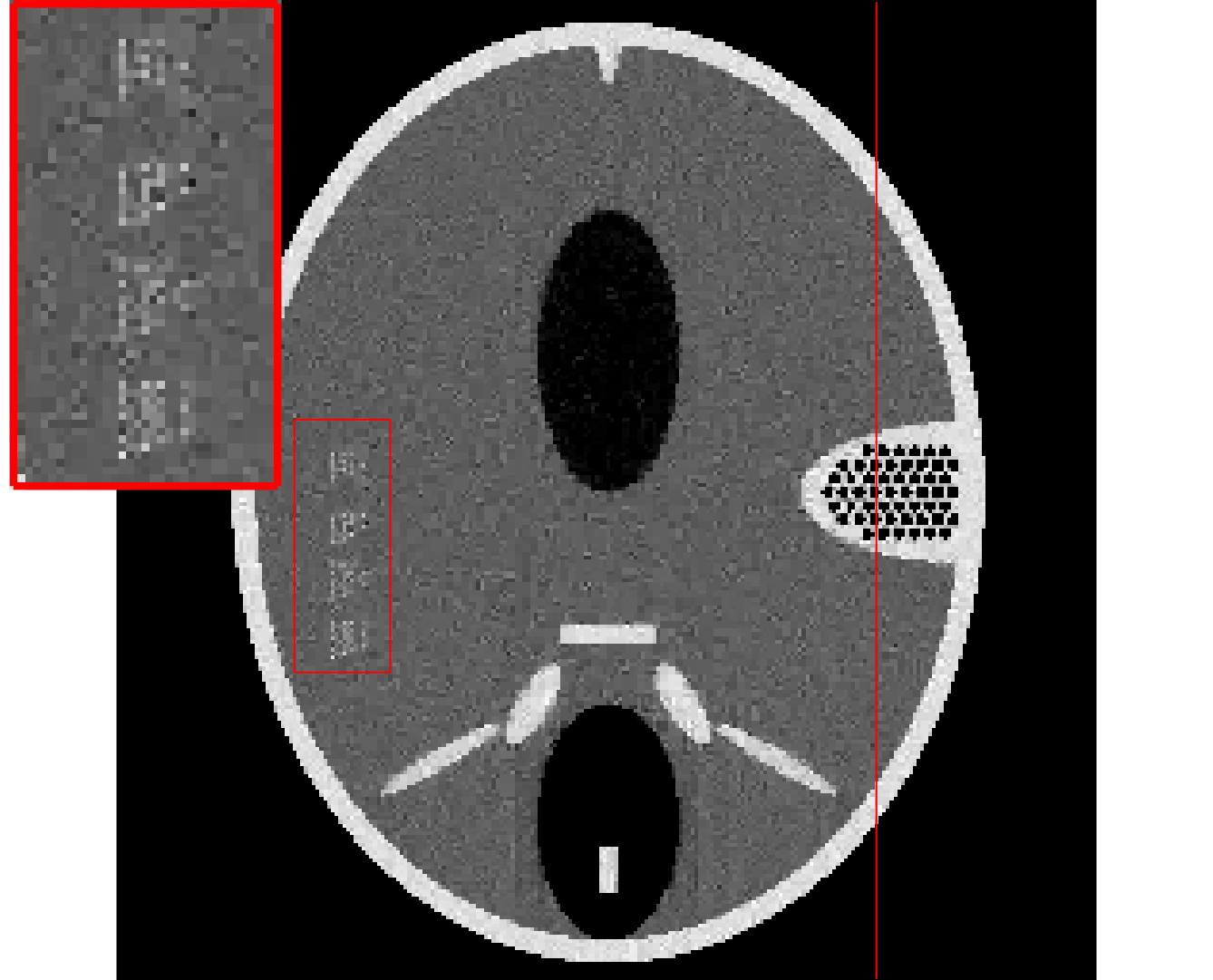}
\captionsetup{justification=centering}
\caption*{(f)}
\end{minipage}%
\begin{minipage}[!h]{0.25\textwidth}
\includegraphics[height=1.4in]{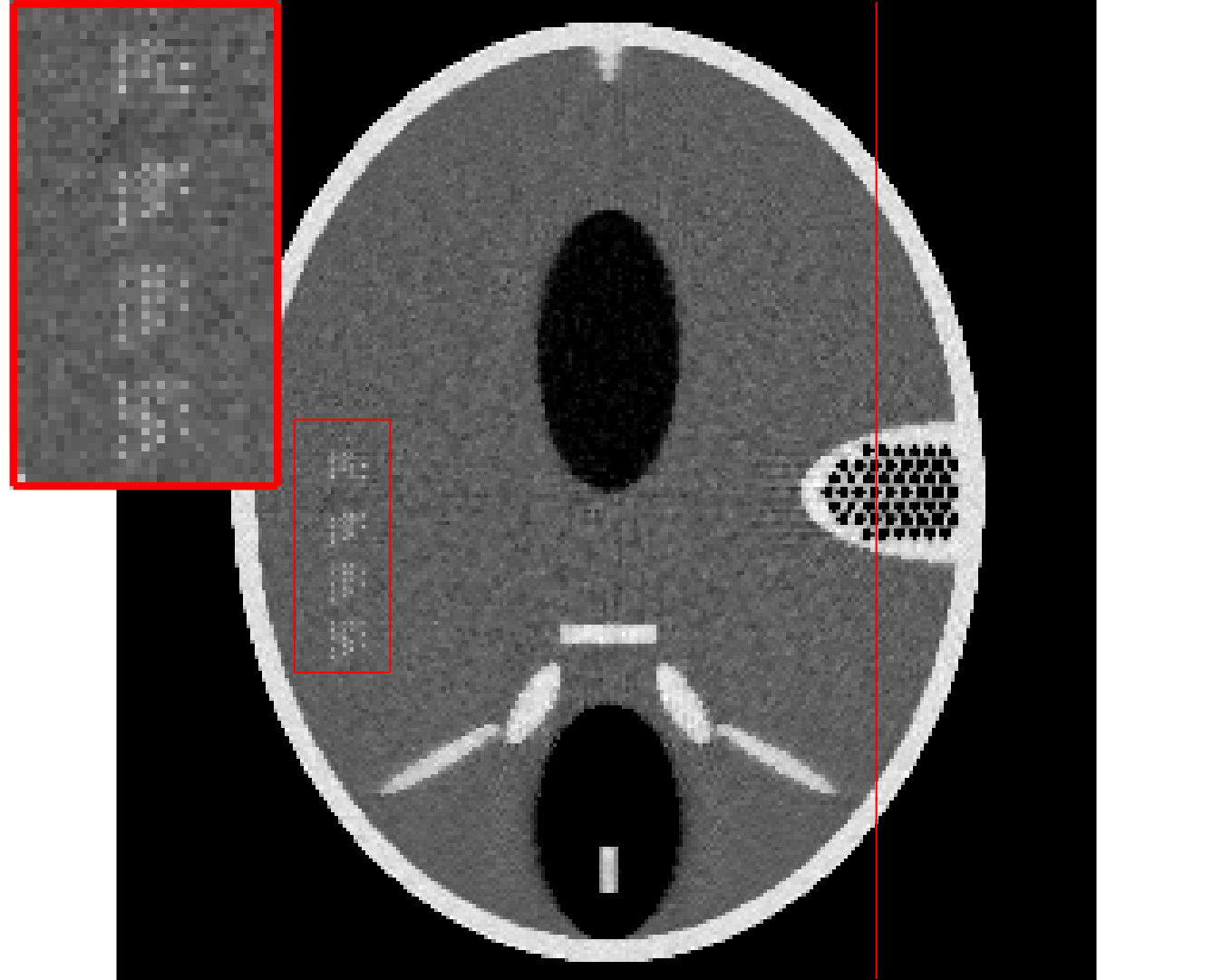}
\captionsetup{justification=centering}
\caption*{(g)}
\end{minipage}%
\begin{minipage}[!h]{0.25\textwidth}
\includegraphics[height=1.4in]{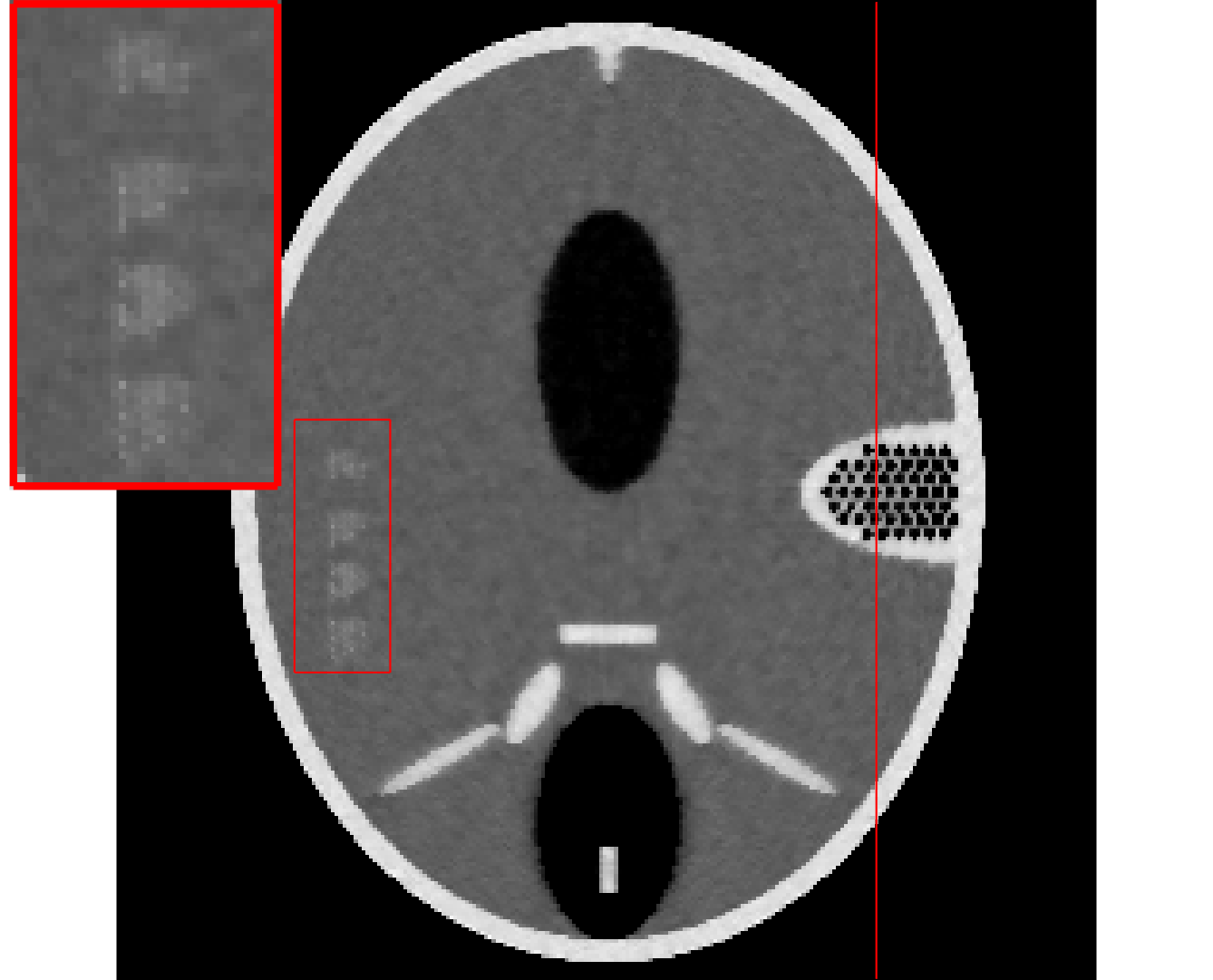}
\captionsetup{justification=centering}
\caption*{(h)}
\end{minipage}%
\caption*{Fig.1 The reconstructed images: (a) ground truth, (b) the Lap-VARD, (c) the traditional VARD, (d) the unpenalized AM, (e) wavelet penalized AM with penalty weight $300$, (f) wavelet penalized AM with penalty weight $1000$, (g) neighborhood penalized AM with penalty weight $2\times 10^5$, (h) neighborhood penalized AM with penalty weight $6\times 10^5$.}
\end{figure*}

\begin{figure*}[!ht]
\begin{minipage}[ht]{0.25\textwidth}
\includegraphics[height=1 in, width =1.8 in]{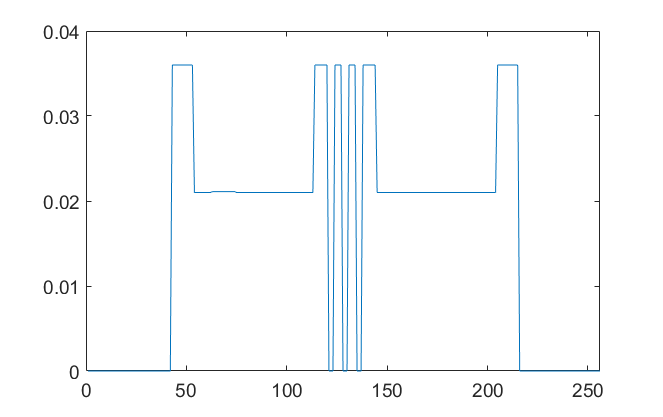}
\caption*{(a)}
\end{minipage}%
\begin{minipage}[ht]{0.25\textwidth}
\includegraphics[height=1in, width =1.8 in]{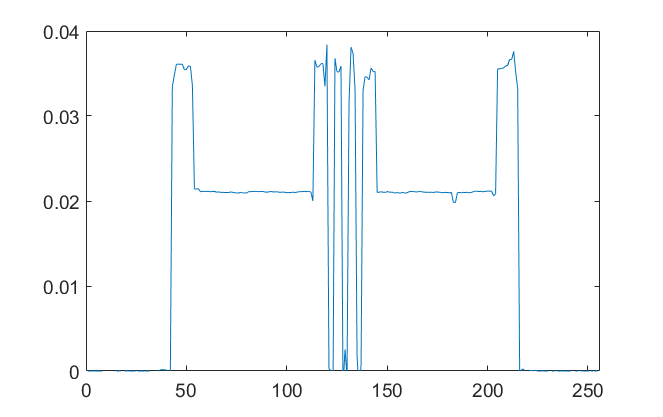}
\caption*{(b)}
\end{minipage}%
\begin{minipage}[ht]{0.25\textwidth}
\includegraphics[height=1 in, width =1.8 in]{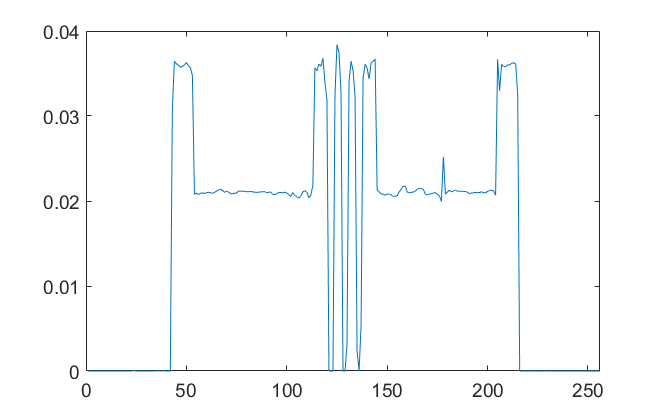}
\caption*{(c)}
\end{minipage}%
\begin{minipage}[ht]{0.25\textwidth}
\includegraphics[height=1 in, width =1.8 in]{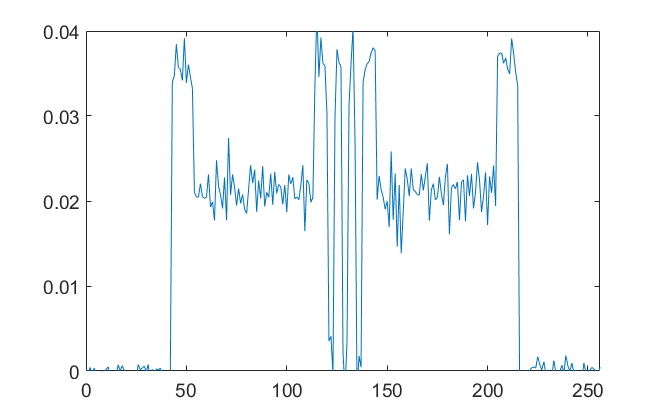}
\caption*{(d)}
\end{minipage}
\begin{minipage}[ht]{0.25\textwidth}
\includegraphics[height=1 in, width =1.8 in]{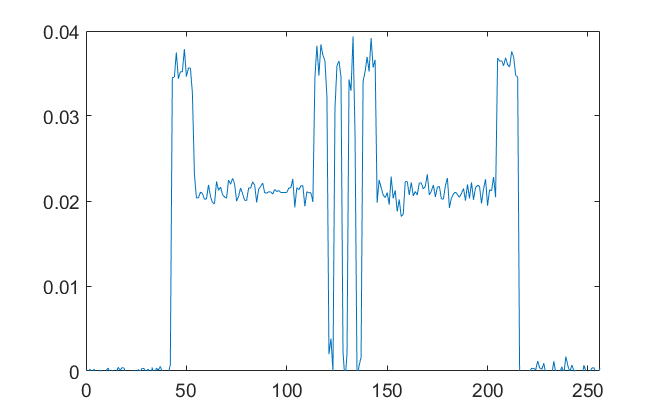}
\caption*{(e)}
\end{minipage}%
\begin{minipage}[ht]{0.25\textwidth}
\includegraphics[height=1 in, width =1.8 in]{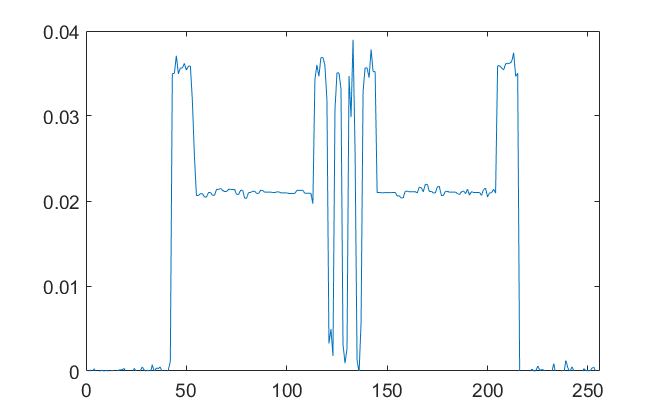}
\caption*{(f)}
\end{minipage}%
\begin{minipage}[ht]{0.25\textwidth}
\includegraphics[height=1 in, width =1.8 in]{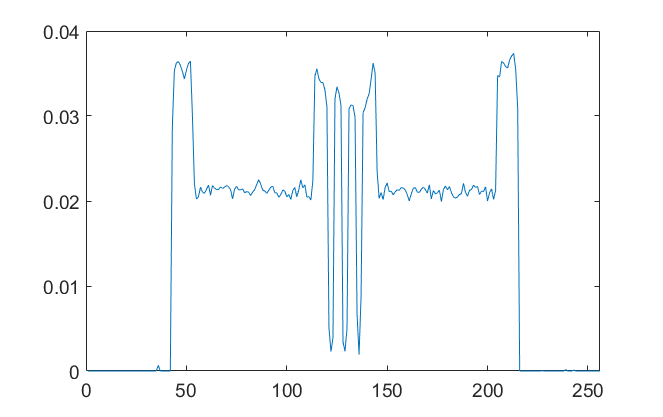}
\caption*{(g)}
\end{minipage}%
\begin{minipage}[ht]{0.25\textwidth}
\includegraphics[height=1 in, width =1.8 in]{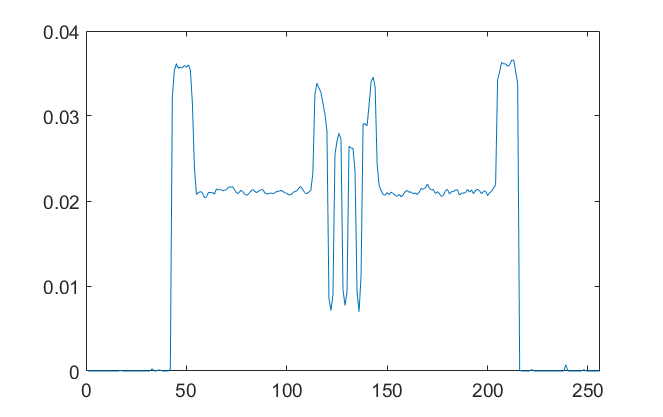}
\caption*{(h)}
\end{minipage}
\caption*{Fig. 2 Profiles of slice 199: (a) ground truth, (b) the Lap-VARD, (c) the traditional VARD, (d) the unpenalized AM, (e) wavelet penalized AM with penalty weight $300$, (f) wavelet penalized AM with penalty weight $1000$, (g) neighborhood penalized AM with penalty weight $2\times 10^5$, (h) neighborhood penalized AM with penalty weight $6\times 10^5$.}
\end{figure*}
\section{Numerical Results}
In this section, we show the performance of Lap-VARD by a phantom simulation experiment. We use the FORBILD head phantom with image size $256\times 256$ \cite{Yu2012}. The original system matrix $H$ has dimension $216000\times 65536$. The Haar wavelet transform with 3 levels is chosen and the corresponding transform matrix $\Omega$ is $65536\times65536$. So the synthesized system matrix $\Phi$ is $216000\times65536$. In this experiment, the photon intensity $I_i=10^5$. We compare the performance of the Lap-VARD with the AM algorithm \cite{OSullivan2007} and penalized AM algorithm with neighborhood penalty \cite{Evans2011} and wavelet sparsity penalty \cite{Xu2014}. The reconstructed images are shown in Fig. 1. The detail structures in the red boxes are magnified on the up-left corners. 

Fig. 1(a) shows the ground truth of a FORBILD phantom. In (b), we plot the inverse wavelet transform of the posterior mean $\mu$ as the reconstructed image; and this image has both low noise and good resolution. The reconstruction of VARD \cite{Kaganovsky2014} is shown in (c); while this image has low noise and good resolution, there are isolated salt-and-pepper noise pixels. Fig. (d) is the AM algorithm reconstruction. Because the Haar wavelet is orthogonal,  the wavelet AM and the traditional AM are equivalent, and the result shows a high noise level. Fig. (e) and (f) are wavelet penalized AM algorithm reconstructions with penalty weights $300$ and $1000$. In (e), the penalty choice is too small and we can still see clear noise  pixels. In (f), we have a proper choice of penalty weight and we have high smoothness and resolution at the same time. Fig (g) and (h) are neighborhood penalized AM reconstructions with penalty weights $2\times 10^5$ and $6\times 10^5$. In (g), we have a proper penalty weight choice with a high resolution and a lower noise level; in (h), the penalty weight is too large and the detail structures are blurred. 

To further compare these algorithms rewording the smoothness and resolution performance, we plot the profiles of the vertical slice No. 199 which are highlighted with red line in Fig. 1. All these profiles are shown in Fig. 2. Fig 2(a) shows the ground truth of the profile. In (b), the reconstruction from the Lap-VARD is shown, we have quantitatively accurate reconstruction in both the flat area and peak-valley contrast. In (c), the profile from the original VARD is plotted, and we see similar performance as the Lap-VARD in smoothness restoration and peak-valley contrast, but we can also see salt-and-pepper noise pixels in VARD. The profile from unpenalized AM is shown in (d), we can see clear noise but the peak and valley contrast is kept. Fig (e) plots the result from wavelet penalized AM algorithm with penalty weight $300$; it still shows high noise compared with the Lap-VARD result which means the weight is insufficient. In (f), with a proper choice of penalty weight $1000$, the wavelet penalized AM result shows that even though it can generate an unbiased result, the noise level is comparatively higher than the Lap-VARD algorithm. Fig (g) is the neighborhood penalized AM result with penalty weight $2\times10^5$, and compared with the Lap-VARD result the noise level is too high. Another severe disadvantage is that the peak and valley values begin to shrink towards the average value.  In an over-smoothed case, as shown in (h), with penalty weight raised to $6\times 10^5$, the peak and valley contrast further shrinks and the result is quantitatively biased.  

A quantitative error comparison is summarized in Table 1. The Lap-VARD outperforms all the other algorithms in root mean square error (RMSE) and peak signal-to-noise ratio (PSNR) performances.
\begin{table*}[!ht]
\centering
\begin{tabular}{cccccccc}
\hline
\hline
&Lap-VARD &VARD& AM &wav-AM ($\gamma=300$) &wav-AM($\gamma=1000$) &AM($\lambda=2\times 10^5$) &AM($\lambda=6\times 10^5$)  \\
\hline
RMSE &$4.86\times 10^{-4}$ &$7.09\times 10^{-4}$&0.0020 &$9.45\times 10^{-4}$ &$7.28\times 10^{-4}$ &0.0012 &0.0010  \\
\hline
PSNR(dB) &38.37 &35.42 &26.74 &32.70 &34.93 &31.79 &29.88  \\ 
\hline
\hline
\end{tabular}
\caption*{Table 1. RMSE and PSNR performances of different algorithms}
\end{table*} 

From the simulation experiment above, we find that the Lap-VARD algorithm is able to derive a optional choice of penalty weight without sacrificing the smoothness or the resolution of reconstructed images.

\section{conclusion}
We introduced the Lap-VARD algorithm and implemented in X-ray computed tomography. The algorithm automatically generates the optimal wavelet penalty weight choice $\gamma$ and the reconstructed image. Compared with a wavelet penalty weight which is from empirical experiments, the input data-driven Lap-VARD algorithm outperforms other algorithms in retaining the smoothness and detailed structure resolution. Compared with using a neighborhood penalty which loses resolution and generates biased results in a high peak and valley contrast scenario, the Lap-VARD is able to keep the contrast and generate an unbiased result.     


%

\ifCLASSOPTIONcaptionsoff
  \newpage
\fi



%
\newpage
\balance
\renewcommand\refname{Reference}
\bibliography{IEEEabrv,SPL_ver}

\begin{thebibliography}{10}
\providecommand{\url}[1]{#1}
\csname url@samestyle\endcsname
\providecommand{\newblock}{\relax}
\providecommand{\bibinfo}[2]{#2}
\providecommand{\BIBentrySTDinterwordspacing}{\spaceskip=0pt\relax}
\providecommand{\BIBentryALTinterwordstretchfactor}{4}
\providecommand{\BIBentryALTinterwordspacing}{\spaceskip=\fontdimen2\font plus
\BIBentryALTinterwordstretchfactor\fontdimen3\font minus
  \fontdimen4\font\relax}
\providecommand{\BIBforeignlanguage}[2]{{%
\expandafter\ifx\csname l@#1\endcsname\relax
\typeout{** WARNING: IEEEtran.bst: No hyphenation pattern has been}%
\typeout{** loaded for the language `#1'. Using the pattern for}%
\typeout{** the default language instead.}%
\else
\language=\csname l@#1\endcsname
\fi
#2}}
\providecommand{\BIBdecl}{\relax}
\BIBdecl

\bibitem{Kalender2006}
W.~A. Kalender, ``X-ray computed tomography,'' \emph{Physics in medicine and
  biology}, vol.~51, no.~13, p. R29, 2006.

\bibitem{Simoncelli1999}
E.~P. Simoncelli, ``Modeling the joint statistics of images in the wavelet
  domain,'' in \emph{SPIE's International Symposium on Optical Science,
  Engineering, and Instrumentation}.\hskip 1em plus 0.5em minus 0.4em\relax
  International Society for Optics and Photonics, 1999, pp. 188--195.

\bibitem{Xu2014}
Q.~Xu, A.~Sawatzky, M.~A. Anastasio, and C.~O. Schirra, ``Sparsity-regularized
  image reconstruction of decomposed k-edge data in spectral ct,''
  \emph{Physics in medicine and biology}, vol.~59, no.~10, p. N65, 2014.

\bibitem{Baraniuk2007}
\BIBentryALTinterwordspacing
R.~Baraniuk, ``Compressive sensing [lecture notes],'' \emph{{IEEE} Signal
  Process. Mag.}, vol.~24, no.~4, pp. 118--121, 2007. [Online]. Available:
  \url{http://ieeexplore.ieee.org/stamp/stamp.jsp?arnumber=4286571}
\BIBentrySTDinterwordspacing

\bibitem{Candes2006}
E.~J. Candes, J.~K. Romberg, and T.~Tao, ``Stable signal recovery from
  incomplete and inaccurate measurements,'' \emph{Communications on pure and
  applied mathematics}, vol.~59, no.~8, pp. 1207--1223, 2006.

\bibitem{Tibshirani1996}
R.~Tibshirani, ``Regression shrinkage and selection via the lasso,''
  \emph{Journal of the Royal Statistical Society. Series B (Methodological)},
  pp. 267--288, 1996.

\bibitem{Zou2006}
H.~Zou, ``The adaptive lasso and its oracle properties,'' \emph{Journal of the
  American statistical association}, vol. 101, no. 476, pp. 1418--1429, 2006.

\bibitem{Hebert1992}
T.~J. Hebert and R.~Leahy, ``Statistic-based map image-reconstruction from
  poisson data using gibbs priors,'' \emph{Signal Processing, IEEE Transactions
  on}, vol.~40, no.~9, pp. 2290--2303, 1992.

\bibitem{Wipf2008}
D.~P. Wipf and S.~S. Nagarajan, ``A new view of automatic relevance
  determination,'' in \emph{Advances in neural information processing systems},
  2008, pp. 1625--1632.

\bibitem{Kaganovsky2014}
Y.~Kaganovsky, S.~Han, S.~Degirmenci, D.~G. Politte, D.~J. Brady, J.~A.
  O'Sullivan, and L.~Carin, ``Alternating minimization algorithm with automatic
  relevance determination for transmission tomography under poisson noise,''
  \emph{arXiv preprint arXiv:1412.8464}, 2014.

\bibitem{Fessler2000}
J.~A. Fessler, ``Statistical image reconstruction methods for transmission
  tomography,'' \emph{Handbook of medical imaging}, vol.~2, pp. 1--70, 2000.

\bibitem{DeMan2001a}
B.~De~Man, J.~Nuyts, P.~Dupont, G.~Marchal, and P.~Suetens, ``An iterative
  maximum-likelihood polychromatic algorithm for ct,'' \emph{IEEE transactions
  on medical imaging}, vol.~20, no.~10, pp. 999--1008, 2001.

\bibitem{Jingwei}
J.~Lu, J.~A. O'Sullivan, and D.~G. Politte, ``Wavelet regularized alternating
  minimization algorithm for low dose x-ray ct,'' in \emph{Proceedings of The
  14th International Meeting on Fully Three-Dimensional Image Reconstruction in
  Radiology and Nuclear Medicine}, 2017, pp. 726--732.

\bibitem{Neal1998}
R.~M. Neal and G.~E. Hinton, ``A view of the em algorithm that justifies
  incremental, sparse, and other variants,'' in \emph{Learning in graphical
  models}.\hskip 1em plus 0.5em minus 0.4em\relax Springer, 1998, pp. 355--368.

\bibitem{Boyd2006}
S.~Boyd and A.~Mutapcic, ``Subgradient methods,'' \emph{Lecture notes of
  EE364b, Stanford University, Winter Quarter}, vol. 2007, 2006.

\bibitem{OSullivan2007}
J.~A. O'Sullivan and J.~Benac, ``Alternating minimization algorithms for
  transmission tomography,'' \emph{Medical Imaging, IEEE Transactions on},
  vol.~26, no.~3, pp. 283--297, 2007.

\bibitem{Yu2012}
Z.~Yu, F.~Noo, F.~Dennerlein, A.~Wunderlich, G.~Lauritsch, and J.~Hornegger,
  ``Simulation tools for two-dimensional experiments in x-ray computed
  tomography using the forbild head phantom,'' \emph{Physics in medicine and
  biology}, vol.~57, no.~13, p. N237, 2012.

\bibitem{Evans2011}
J.~D. Evans, D.~G. Politte, B.~R. Whiting, J.~A. O'Sullivan, and J.~F.
  Williamson, ``Noise-resolution tradeoffs in x-ray ct imaging: A comparison of
  penalized alternating minimization and filtered backprojection algorithms,''
  \emph{Medical physics}, vol.~38, no.~3, pp. 1444--1458, 2011.

\end{thebibliography}

%








\end{document}